\pdfoutput=1

\documentclass[11pt]{article}

\usepackage[final]{acl}

\usepackage{times}
\usepackage{latexsym}

\usepackage[T1]{fontenc}

\usepackage[utf8]{inputenc}

\usepackage{microtype}

\usepackage{inconsolata}

\usepackage{graphicx}
\usepackage{booktabs}
\usepackage{xcolor}
\usepackage{comment}

\title{
Contextual morphologically-guided tokenization for Latin encoder models
}

\author{
  \textbf{Marisa Hudspeth\textsuperscript{1}} \quad
  \textbf{Patrick J. Burns\textsuperscript{2}} \quad
  \textbf{Brendan O'Connor\textsuperscript{1}}
\\
\\
  \textsuperscript{1}Manning College of Information \& Computer Sciences, University of Massachusetts Amherst \\
  \textsuperscript{2}Institute for the Study of the Ancient World, New York University
\\
    \texttt{\{mhudspeth,brenocon\}@cs.umass.edu} \quad
    \texttt{pjb311@nyu.edu}
}

\begin{document}
\maketitle
\begin{abstract}
Tokenization is a critical component of
language model pretraining, yet standard tokenization methods often prioritize information-theoretical goals like high compression and low fertility rather than linguistic goals like morphological alignment.
In fact,
they have been shown to be suboptimal for morphologically rich languages, where tokenization quality directly impacts downstream performance. 
In this work, we investigate morphologically-aware tokenization for Latin, a morphologically rich language that is medium-resource in terms of pretraining data, but high-resource in terms of curated lexical resources -- a distinction that is often overlooked but critical in discussions of low-resource language modeling.
We find that morphologically-guided tokenization improves overall performance on four downstream tasks. Performance gains are most pronounced for out of domain texts, highlighting our models' improved generalization ability.
Our findings demonstrate the utility of linguistic resources to improve language modeling for morphologically complex languages. For low-resource languages that lack large-scale pretraining data, the development and incorporation of linguistic resources can serve as a feasible alternative to improve LM performance.\footnote{Code and data are available here: \url{https://github.com/slanglab/latin-morpheme-tokenization}}
\end{abstract}

\section{Introduction}
Tokenization is the first step in Large Language Model (LLM) 
pretraining pipeline, making it the foundation upon which model performance rests. A common assumption is that tokenizers should maximize compression and minimize fertility \cite{Schmidt2024TokenizationIM}. However, recent research has challenged this view, particularly in the context of morphologically rich and lower-resource languages. Studies have shown that existing tokenization methods exhibit low morphological alignment \cite{Hsu2023MorphPieceA,bostrom-durrett-2020-byte, turkish-thesis,Libovicky2024LexicallyGS}, 
which can negatively impact downstream performance. In this work, we investigate tokenization strategies for Latin, a morphologically rich, medium-resource language\footnote{Between 100M and 1B tokens \cite{chang-etal-2024-multilinguality}} with a long scholarly tradition, and moreover one with specific interest in word endings and other aspects of word formation, 
making it an informative test case.

We hypothesize that incorporating morphological knowledge into tokenization will improve both morphological alignment and downstream performance. While prior work has explored morphologically-aware tokenizers, they often focus on high-resource and/or morphologically simple languages \cite{hofmann-etal-2021-superbizarre,Hsu2023MorphPieceA,bostrom-durrett-2020-byte} or employ acontextual, unsupervised morphological analyzers such as Morfessor \cite{morfessor}. Furthermore, evaluations in this area have not examined fine-grained morphological feature prediction, which should better capture whether a morphologically-aligned tokenization helps the language model's contextual embeddings capture its linguistic content.\footnote{Much prior work attains a partial view of this by evaluating on POS tagging, a significantly coarser version of the problem.}


Beyond its computational implications, this question is of particular interest within Latin linguistic and philological research. The role of word endings in meaning construction has been central to Latin scholarship for centuries, making it crucial to empirically test whether morphology-informed tokenization aligns with these long-standing linguistic intuitions.

One of the unique advantages of working with Latin is the availability of expert-curated morphological and lexical resources---benefiting from long-standing philological study about the role of word endings in constructing meaning \cite{diederich-latin-endings, pellegrini2021two}.  
Although Latin lacks the raw text data suitable for modern, large-scale pretraining, this abundance of linguistic resources makes Latin
a useful case study to investigate whether the development of comparable resources for low-resource languages can meaningfully improve language modeling performance, especially in cases where acquiring more pretraining data is infeasible.



To test our hypothesis, we experiment with three types of tokenizers:
\begin{enumerate}
    \vspace{-2mm}
    \item \textbf{Baseline:} Standard WordPiece (WP) and Unigram Language Model (ULM)
    \vspace{-2mm}
    \item \textbf{Low-guidance approach:} Seeding morphological suffixes into the tokenizer vocabulary. This approach is lightweight, only requiring a predefined list of suffixes.
    \vspace{-2mm}
    \item \textbf{High-guidance approach:} Pre-Tokenization using a morphological analyzer. Unlike prior work, we disambiguate the analyses based on POS information, making our method context-aware. 
    \vspace{-1mm}
\end{enumerate}

\noindent

For evaluation, we pretrain Latin RoBERTa models \cite{liu2019robertarobustlyoptimizedbert} using each tokenizer and test them on \textbf{four downstream tasks}: POS and morphological feature tagging, named entity recognition (NER), word sense disambiguation (WSD), and authorship verification (AV). To our knowledge, no prior studies have evaluated their morphologically aware tokenization methods using morphological feature classification as a downstream task. 


\section{Related Work}
\subsection{Tokenization}\label{sec:rel_work_tokenization}
Tokenization plays a crucial role in language model pretraining, yet its impact on morphologically rich languages remains an active area of investigation. Studies have increasingly questioned whether widely used tokenization methods such as Byte Pair Encoding (BPE) \cite{sennrich-etal-2016-neural}, Unigram Language Model (ULM) \cite{kudo-2018-subword}, and WordPiece \cite{Schuster2012JapaneseAK} sufficiently capture morphological structure, and whether this matters for downstream performance.

Several studies suggest that aligning BPE and ULM tokenizers' segmentations with gold-standard morphological boundaries can enhance downstream performance on sentiment and topic classification (\citet{hofmann-etal-2021-superbizarre}; English), zero-shot summarization and retrieval (\citet{Hsu2023MorphPieceA}; English), QA, MNLI, NER (\citet{bostrom-durrett-2020-byte}; English and Japanese), and other classification, structured prediction, and similarity tasks (\citet{Vemula2025RethinkingTF}; English, Hindi, and Telugu).

Studies on morphologically rich languages in particular provide further evidence for the benefits of morphology-aware tokenization. Rule-based approaches have shown improvements in Romanian NLP tasks \cite{Vasiu2020EnhancingTB}, and pre-tokenization methods using either 1) a language-specific morphological analyzer \cite{turkish-thesis,nzeyimana-niyongabo-rubungo-2022-kinyabert,Vemula2025RethinkingTF}, or 2) unsupervised methods such as Morfessor \cite{park-etal-2021-morphology,Libovicky2024LexicallyGS, Vemula2025RethinkingTF} have all yielded downstream performance gains. Post-training strategies have also been effective; for instance, modifying existing BPE vocabularies improved token-level tasks in English, Dutch, and German \cite{bauwens-delobelle-2024-bpe}.

While most morphologically-aware tokenization methods rely on static rules or unsupervised segmentation, some studies have experimented with adding contextual information. \citet{yehezkel-pinter-2023-incorporating} introduced SaGe, a tokenizer whose vocabulary construction method closely resembles ULM but incorporates a SkipGram objective to refine vocabulary selection. This approach improved performance on English GLUE and NER, and Turkish Inference and NER.


Other studies directly oppose the hypothesis that morphological alignment is beneficial. \citet{Toraman2022ImpactOT} found no improvements in Turkish NLP tasks when pre-tokenizing with a morphological analyzer, though they noted that errors in the analyzer itself may have influenced results. More broadly, \citet{arnett-bergen-2025-language} argued that morphological alignment is not a key factor in tokenization quality, emphasizing instead that dataset size and quality are more important. \citet{arnett2025evaluating} even find a small negative correlation between morphological alignment and downstream task performance across 70 languages. However, this analysis only considers previously reported measures of downstream task performance when they exist--which is typically only for high-resource languages, and for large multilingual models--and perplexity when they do not (understandably, since the very nature of low resource languages means they lack evaluation data). 

The existing literature provides strong, though not unanimous, evidence that morphologically-aware tokenization can improve NLP performance, particularly for morphologically rich languages. However, prior studies have largely focused on a limited set of downstream tasks---such as POS tagging and NER---which may not fully expose the benefits of morphologically-aligned tokenization. Our work extends this research by evaluating multiple tokenization strategies for Latin, including both light and high-guidance approaches. Additionally, we introduce morphological feature classification as an alternative downstream evaluation metric, hypothesizing that the fine-grained morphological feature values will reveal improvements that are less evident in coarse-grained POS tagging.




\subsection{Morpheme Segmentation}


Morpheme segmentation has been widely studied as an independent NLP task, distinct from its potential applications in tokenization and language model pretraining. One of the most prominent efforts in this area is the SIGMORPHON shared task on morpheme segmentation \cite{batsuren-etal-2022-sigmorphon}, which provided segmentation data for nine languages, including Latin. This task included both acontextual and contextual segmentation challenges; however, Latin was excluded from the contextual segmentation track. While the availability of segmentation resources for these nine languages is valuable, the dataset is too small to support pretraining efforts. Moreover, many morphological datasets, including those used in SIGMORPHON, are constructed through automatic extraction from sources such as Wiktionary, introducing data quality concerns. For example, \citet{gorman-etal-2019-weird} highlight extensive extraction errors in the dataset for the 2017 CoNLL-SIGMORPHON shared task for Morphological Reinflection \cite{cotterell-etal-2017-conll}.

Latin is unique within morpheme segmentation research due to its rich morphological tradition and availability of high-quality, expert-curated resources. Unlike many other languages, Latin benefits from centuries of linguistic study focused on word formation and morphological structure \cite{diederich-latin-endings, pellegrini2021two}.

Our work builds on the precisely curated morphological resources available for Latin, incorporating linguistic knowledge at the morpheme level in a way that is not feasible for many other languages. We argue that context-aware morphological tokenization--though requiring significant language-specific effort--has the potential to bridge the gap between linguistic theory and modern NLP. 

\subsection{Language Modeling}
Several studies have demonstrated the feasibility of pretraining transformer-based models for low-resource languages. \citet{ogueji-etal-2021-small} introduced AfriBERTa, a multilingual BERT model trained on various low-resource African languages using a corpus of approximately 1GB, comparable in size to ours. Their model outperforms massively multilingual models like mBERT and XLM-R \cite{conneau-etal-2020-unsupervised} on NER and text classification. Similarly, \citet{chang-etal-2024-multilinguality} systematically evaluated the relationship between model size (from tiny to small) and data size across both monolingual and multilingual GPT-2 models \cite{Radford2019LanguageMA}. 


Prior work in Latin language modeling has produced several pretrained, transformer-based Latin models.
LaBERTa \cite{riemenschneider-frank-2023-exploring} was trained on 165M words from Corpus Corporum \cite{Roelli2014TheCC},\footnote{\url{https://mlat.uzh.ch/}} a kind of super-repository of available smaller digitized Latin text repositories. 
Another Latin RoBERTa model \cite{stroebel-roberta-base-latin-cased1, liu2019robertarobustlyoptimizedbert} was trained on a 390M token corpus also derived from Corpus Corporum.
Finally, LatinBERT \cite{bamman2020latinbertcontextuallanguage} was trained on a larger corpus (642.7M words), though a significant portion originated from noisy OCR-processed Latin texts from the Internet Archive. Its cleaner subset contained 81.6M words.

Our work differs from these prior efforts in Latin language modeling in two ways. 
First, our pretraining training corpus, totaling 195M words (1GB), is larger than the clean subset used in LatinBERT though smaller than the dataset used for Latin RoBERTa. 
Finally, we experiment with various morphologically-aware tokenization strategies, whereas existing Latin language models use baseline WordPiece (LatinBERT) and BPE (LaBERTa, Latin RoBERTa)..


\section{Background: Tokenization}
\citet{Schmidt2024TokenizationIM} conceptualize tokenization as a three-step process: 1) pretokenization, which applies an initial set of rules to define processing units--typically by segmenting on whitespace; 2) vocabulary construction or training, where subword units are learned; 3) segmentation or decoding, which determines how input text is tokenized based on the trained vocabulary.
This framework helps highlight the different places where morphological guidance can be introduced, instead of viewing tokenizers as indivisible systems.
We experiment with modifications to two widely-used Huggingface tokenizer implementations.


\subsection{WordPiece Tokenization}
BPE and WordPiece are tokenizers with similar training algorithms. While BPE is widely used, prior work finds it suboptimal (\S\ref{sec:rel_work_tokenization}), so our experiments focus on WordPiece. For clarity, we overview both in this section.

Pretokenization for both tokenizers is typically done by splitting on whitespace and punctuation. 
Given a list of (pretokenized) strings and a desired final vocabulary size, an initial subword vocabulary is constructed from all unique characters. Subword types are iteratively merged until the desired vocabulary size is reached. For WordPiece, the subword bigram with the highest pointwise mutual information (PMI) \cite{bouma2009normalized} is chosen. Its two subwords are merged into a new, single subword and added to the vocabulary. For BPE, the bigram with the highest frequency is chosen rather than highest PMI.

To tokenize new text, WordPiece performs greedy left-to-right decoding, whereas BPE applies merge rules in the order learned during training. 

\subsection{Unigram Language Model Tokenization}
The Unigram Language Model \cite{kudo-richardson-2018-sentencepiece} infers the most likely segmentation for a word, using the Viterbi algorithm. For learning, after initial pretokenization,\footnote{Pretokenization is also done by splitting on whitespace. In the HuggingFace implementation, punctuation is not split on.}
the model starts with a large vocabulary of all substrings in the corpus. Subwords are iteratively pruned in order to maximize the unigram likelihood of the corpus until the desired vocabulary size is reached.


\section{Method} \label{sec:method}

\subsection{Morphologically-Enhanced Tokenizers} \label{sec:morph_enhanced_tokenizers}
We add morphological guidance to both ULM and WordPiece tokenizer models, implemented by modifying HuggingFace's implementations of each \cite{wolf-etal-2020-transformers}.
We create and evaluate three tokenizer variations: MorphSeeding, MorphPreTokenization (acontextual), and MorphPreTokenization (contextual).

\paragraph{Data} We train our tokenizers on our pretraining corpus.\footnote{The ULM tokenizers are trained on 5\% of this corpus. See \S\ref{sec:limitations}.}

\paragraph{MorphSeeding} We create a list of 480 morphological suffixes sourced from Lemlat, a type-level lemmatizer and morphological analyzer for Latin \cite{passarotti-etal-2017-lemlat}. For our purposes, 
we define morphological suffixes as all segments of a word which are not the first (root/stem) segment.

Then, we modify the ULM and WordPiece trainers to bias them to prefer segmenting with this list of suffixes. For WordPiece, all suffixes are added to the initial vocabulary with the continuing subword prefix "\#\#" prepended. Since WordPiece's vocabulary construction is bottom-up, once added to the vocabulary a subword cannot be removed. For ULM, all suffixes are added to the initial vocabulary, and for decoding, their log-probabilities are upweighted by a fixed amount\footnote{During initial experiments, we found a weight of 0.5 to strike a good balance of encouraging these suffixes to be chosen more often when decoding, while not increasing fertility to an unreasonable degree.} in the lattice. Suffixes are not allowed to be removed from the vocabulary.

\paragraph{MorphPreTokenization} \label{sec:morph_pretokenization_method}
We analyze all unique words in our corpus with Lemlat.
For each word, Lemlat returns a list of possible analyses that include the word's segmentation into morphemes, 
as well as its lemma, declension or conjugation, part of speech, morphological features, and derivational affixes. We only utilize the segmentation and POS.

We then presegment these morphemes in our corpus, so that during tokenizer training, it will never merge Lemlat-provided morphemes.
We experiment with both contextual and acontextual segmenters.

For acontextual pretokenization, we simply use the segmentation in the first analysis given by Lemlat. This follows the type-level focus of previous work, either with language-specific morphological analyzers \cite{Toraman2022ImpactOT, turkish-thesis, nzeyimana-niyongabo-rubungo-2022-kinyabert} or the unsupervised Morfessor model \cite{morfessor, Libovicky2024LexicallyGS}.

But in many instances, there exists ambiguity over a word's segmentation, which can be resolved with contextual information about its grammatical role.  Thus, we construct a contextual morphological segmenter by first running an off-the-shelf part-of-speech tagger on the corpus, and filtering Lemlat's output to an analysis with a matching POS tag.\footnote{Interestingly, for some tasks the approach may seem circular: a predicted POS tag helps guide the LLM tokenization, and thus the eventual LLM contextual representation used to predict, for example, a POS tag. Investigating how initial tagging errors propagate would be interesting future work.}

We tag our corpus with LatinCy \cite{burns2023latincysynthetictrainedpipelines}. It uses the Latin UD Treebanks' tagset, which differs from Lemlat's. We create a mapping between the tag systems (Table \ref{tab:ud_to_lemlat}), and a protocol for selecting a word's segmentation when the POS tags do not match:
\begin{itemize}
    \vspace{-2mm}
    \item If Lemlat only gives one unique possible segmentation, use that one (occurs in 1.2\% of words in UD treebanks).
    \vspace{-3mm}
    \item If Lemlat gives multiple possible segmentations but none match the predicted POS, do not segment the word (occurs in 0.028\% of words in UD treebanks).
    \vspace{-2mm}
\end{itemize}

\noindent
A word's tag usually disambiguates the segmentation, but in
rare cases, one word may have multiple analyses with the same POS tag, due to multiple possible lemmas or morphological features.
In these cases, we choose one segmentation based on the following criteria: 
\begin{itemize}
    \vspace{-2mm}
    \item If the candidate segmentations have the same number of subwords, choose the one with the longer suffix (i.e. out of the adjective [\textit{adversar}, \textit{-i}] versus infinitive verb [\textit{advers}, \textit{-ari}] choose the latter).
    \vspace{-3mm}
    \item If the candidate segmentations have a different number of subwords, choose the one with more subwords (i.e. out of participle [\textit{inordin}, \textit{-at}, \textit{-o}] and imperative [\textit{inordin}, \textit{-ato}], choose the former).
    \vspace{-2mm}
\end{itemize}
This type of conflict occurred in 4.55\% of all Lemlat analyses of unique words in the Latin treebanks.\footnote{In both the UD treebanks and in LASLA \cite{denooz2004opera-lasla}, a non-UD Latin treebank.}

This results in four morphoglical pretokenization-based tokenizers---for each model class (ULM and WordPiece), there is both acontextual and contextual presegmentation.  

We implement changes to the tokenizers to accommodate presegmentation.  For ULM, the only modification is to add a new pre-tokenization step which splits on our morpheme symbol, allowing morphological suffixes to be treated as continuing subwords. Due to how the ULM vocabulary is constructed, it is possible that the suffixes will be split into multiple subwords, just like the root.

WordPiece requires modification to its trainer, not just the pretokenizer; for implementation details, see \S\ref{sec:appendix_tok_details}. Unlike ULM, once the suffix subword is added to the WordPiece vocabulary, it will remain unchanged, neither split or merged.

\subsection{Tokenizer Evaluation}
Several metrics have been proposed to assess tokenizer quality.

Renyi entropy \cite{zouhar-etal-2023-tokenization} measures the uniformity of token frequency distributions. However, \citet{Schmidt2024TokenizationIM} found that it correlates with Corpus Token Count, which they argue is a poor predictor of downstream performance.

A more linguistically motivated approach is morphological alignment with a gold reference segmentation, which assumes that having "meaningful" subword units improves downstream task performance. Various metrics have been introduced to quantify this.

MorphScore \cite{arnett-bergen-2025-language,arnett2025evaluating} assigns a score of 1 if a tokenizer correctly segments at a specific morpheme boundary, regardless of other boundaries in the word, and 0 otherwise. Unlike other measures, it excludes words that remain unsegmented.

Suffix precision, recall, and f1 \cite{turkish-thesis} evaluate how well a tokenizer captures \textit{suffix} boundaries specifically. 
Subword boundary precision, recall, f1 \cite{bostrom-durrett-2020-byte} which we adopt in this work, assess overall segmentation accuracy. In addition, we track exact matches between predicted and gold segmentations. 

The reliability of these metrics depends on the quality of the gold standard segmentations. Many studies experiment with multiple languages and rely on morphological data scraped from Wiktionary. \citet{gorman-etal-2019-weird} highlight extensive errors in SIGMORPHON’s morpheme reinflection data \cite{cotterell-etal-2017-conll}, demonstrating that such resources may introduce inconsistencies. This underscores the importance of carefully curating high-quality gold standards, 
which tends to be easier when focusing on a single language. When Latin scholar coauthors reviewed the SIGMORPHON segmentation dictionary alongside two open-source Latin morphological dictionaries (Lemlat and Whitaker's Words\footnote{\url{https://latin-words.com/}}), we judged Lemlat to be highest quality. 
Lemlat has also been shown to have better coverage of Latin word types and tokens than Words, and equivalent coverage to LatMor, a finite state transducer for Latin \cite{latmor}.
\footnote{Lemlat does not include macrons, the accent marks that indicate vowel length in Latin (e.g., mālum ‘apple’ vs. malum ‘evil’). While useful for phonological, morphological, or metrical analysis, macrons are an editorial choice, primarily used in poetry or educational material like textbooks and dictionaries. Most regular Latin texts do not include them, and adding them automatically can introduce errors.}

To construct an evaluation set, we extract all unique (word, POS) pairs from the five Latin UD test sets, and segment them using Lemlat. In the acontextual setting, we ignore POS and consider the first segmentation given by Lemlat as the gold segmentation. In the contextual setting, we disambiguate Lemlat's analyses using the gold UD POS, choosing the gold segmentation as described in \S\ref{sec:morph_pretokenization_method}.  

We also test the morphological alignment of our tokenizers as compared to Latin SIGMORPHON 2022 segmentations, which are acontextual \cite{batsuren-etal-2022-sigmorphon}. This provides a fairer evaluation than testing against Lemlat segmentations, since we intentionally design our tokenizers to incorporate morphological information from Lemlat.


\subsection{Model Pretraining}
\label{sec:datamodels}
\paragraph{Data} We train our tokenizers and pretrain our RoBERTa models on the same data used to train the static floret vectors \cite{Boyd_Warmerdam_2022} used in LatinCy, 
a spaCy  pipeline for Latin
\cite{burns2023latincysynthetictrainedpipelines,spacy2}. It is 1.08GB, containing 13.5M sentences
and 195M whitespace-separated words.
This is comparable to the pretraining data size of other Latin transformer models; for example, \citet{riemenschneider-frank-2023-exploring} trained LaBERTa on 167.5M words.

\paragraph{Models} We pretrain eight base (110M parameter) RoBERTa models \cite{liu2019robertarobustlyoptimizedbert} using the HuggingFace Transformers library \cite{wolf-etal-2020-transformers}. 
See \S\ref{sec:appendix_pretrain_details} for details on hyperparameters. 

\subsection{Downstream Tasks}
\begin{table}[h!]
    \centering
    \small
    \setlength{\tabcolsep}{4pt}
    \begin{tabular}{l l l}
        \toprule
        \textbf{Task} & \textbf{Test Size} & \textbf{Paper / Source} \\
        \midrule
        Morph & 92{,}517 words & \citet{hudspeth-etal-2024-latin} \\
        NER & 1{,}618 BI-labels & \citet{beersmans-etal-2023-training} \\
        WSD & 40 lemmas & \citet{ghinassi-etal-2024-language} \\
        AV & 220 text pairs & \citet{gorovaia-etal-2024-sui} \\
        \bottomrule
    \end{tabular}
    \caption{Scale and sources of downstream task evaluation sets.}
    \label{tab:task_summary}
\end{table}
We evaluate our models on four downstream tasks: two token-level (POS/Morphological feature classification, NER) and two sequence-level (WSD and Authorship Verification (AV)). We report overall test set sizes in Table \ref{tab:task_summary}, and sizes of in and out domain splits in Table \ref{tab:testset_sizes_inout}. More detailed descriptions of each task can be found in \ref{sec:app_task_desc}.

For each task, we report the following metrics:
whole-string morphological accuracy and per-feature macro F1 for POS and morphological feature classification; BI-label F1 and per-entity micro F1 for NER; average F1 over lemmas for word sense disambiguation (WSD); and average F1 across trials for authorship verification (AV).

We do not directly compare our models to existing Latin encoder models such as LatinBERT or LaBERTa, as their training data and setup differ and are not publicly available. Our experiments aim to isolate the impact of tokenization strategies on downstream performance, so comparisons are limited to models trained under our controlled conditions. See \ref{sec:app_existing_models} for further discussion.

\section{Results}
\subsection{Tokenizer Evaluation}

\setlength{\tabcolsep}{3pt}
\begin{table}[h]
\vspace{-2mm}
    \centering
    \small
    \setlength{\tabcolsep}{3pt}
    \begin{tabular}{ll|cc|cc}
        \toprule
        \textbf{Tokenizer Type} & \textbf{Model} &
        \multicolumn{2}{c|}{\textbf{Sig (Actx)}} &
        \multicolumn{2}{c}{\textbf{Lem (Ctx)}} \\
         & \textbf{Class} &
        \textbf{EM} & \textbf{Fert.} &
        \textbf{EM} & \textbf{Fert.} \\
        \midrule
        Baseline & ULM & 7.19 & 3.10 & 10.76 & 1.87 \\
        MorphSeed & ULM & 7.03 & 3.09 & 12.12 & 1.89 \\
        MorphPreTok Actx & ULM & \textbf{9.89} & 3.19 & 65.05 & 2.36 \\
        MorphPreTok Ctx & ULM & 9.34 & 3.20 & 71.88 & 2.41 \\
        \hline
        Baseline & WP & 3.60 & 2.84 & 20.12 & 1.77 \\
        MorphSeed & WP & 3.46 & 2.76 & 20.18 & 1.77 \\
        MorphPreTok Actx & WP & 8.45 & 2.67 & 75.50 & 2.13 \\
        MorphPreTok Ctx & WP & 8.28 & 2.65 & \textbf{84.32} & 2.18 \\
        \bottomrule
    \end{tabular}
    \caption{Tokenizers’ morphological segmentation accuracy across acontextual (Sigmorphon test set) and contextual (Lemlat) settings, in terms of Exact Match rate (EM) and Fertility (Fert.). Gold fertility is 2.49 for Sigmorphon and 1.94 for Lemlat.}
    \label{tab:exact_match_small}
\vspace{-2mm}
\end{table}

We evaluate the morphological alignment of each tokenizer by comparing predicted segmentations to a gold-standard segmentations from Lemlat and from Sigmorphon 2022. As seen in Table \ref{tab:exact_match_small}, baseline WordPiece already exhibits relatively strong alignment with gold Lemlat segmentations, outperforming ULM in this regard (+9.4\%  exact match against the gold contextual segmentations). However, baseline ULM is more closely aligned to the Sigmorphon segmentations than baseline WordPiece. This may be due to baseline ULM's higher fertility.
\footnote{For complete evaluation metrics, see Table \ref{tab:tokenizer_eval}.}

Introducing morphological pretokenization (MorphPreTok) significantly enhances alignment to Lemlat for both ULM and WordPiece, with exact matches exceeding 65\% for all variants. Alignment to Sigmorphon also increases between 3-5\%. Although the exact match rate is low compared to Lemlat, this is expected since we intentionally design our tokenizers to incorporate morphological information from Lemlat.
These results suggests that explicitly incorporating morphological information during pretokenization leads to segmentations that better adhere to linguistic ground truth.

By contrast, morphological suffix seeding (MorphSeed) provides only a modest improvement for ULM (+1.4\% exact match against contextual Lemlat segmentations), while having no effect on WordPiece’s alignment. Alignment to Sigmorphon is roughly the same for both ULM and WordPiece. This suggests that while suffix seeding can nudge segmentation toward morphological boundaries, they are less effective than full pretokenization in enforcing linguistically coherent segmentations.

This aligns with previous findings that pretokenization has a larger impact on the resulting vocabulary than the tokenization algorithm itself, since pretokenization places a hard constraint on the maximum token length \cite{velayuthan-sarveswaran-2025-egalitarian}.

\subsection{Downstream Performance}
\setlength{\tabcolsep}{3pt}
\begin{table}[h!]
    \centering
    \small
    
    \begin{tabular}{l l | c c c c}
        \hline
        \textbf{Model} & \textbf{Tok} & \textbf{Morph} & \textbf{NER} & \textbf{WSD} & \textbf{AV} \\
        \midrule
        Baseline & ULM & 89.46 & 65.89 & 61.83 & 62.16 \\
        MorphSeed & ULM & 89.51 & 66.54 & 61.11 & 64.52 \\
        MorphPreTok Actx & ULM & 90.98 & \textbf{73.52} & 59.60 & 65.21 \\
        MorphPreTok Ctx & ULM & 91.00 & 73.07 & 57.73 & \textbf{67.23} \\
        \hline
        Baseline & WP & 89.86 & 66.15 & 57.93 & 66.80 \\
        MorphSeed & WP & 89.82 & 67.72 & 60.75 & 65.07 \\
        MorphPreTok Actx & WP & 91.09 & 69.47 & 56.83 & 61.64 \\
        MorphPreTok Ctx & WP & \textbf{91.18} & 72.72 & \textbf{63.93} & 64.92 \\
        \hline
    \end{tabular}
    \caption{Summary of results on downstream tasks. For Morph classification and NER, we report results using the last subword token for prediction.}
    \label{tab:results_summary}
\end{table}

\paragraph{Overall improvement across all tasks} As reported in Table \ref{tab:results_summary}, all tasks benefit from increased morphological guidance. For morphological feature tagging and NER, the MorphPreTok tokenization consistently improves performance for both the ULM (+1.5 morph acc, +7.6 NER BI f1) and WP (+1.3 morph acc, +6.6 NER BI f1) models. The MorphSeed method also shows minor gains for NER (+1.6 BI f1), but otherwise performs similarly to baselines.

Results are more mixed for WSD and Authorship Verification (AV). For the ULM models, WSD performance is hurt with more morphological guidance (-4.1 f1), but AV performance is improved (+5.1 avg f1). The reverse is true for the WP models (+6.0 WSD f1, -5.2 AV avg f1).

\paragraph{Larger gains for out of domain texts} 
\setlength{\tabcolsep}{4pt}
\begin{table}[h!]
\centering
\small
\begin{tabular}{l l | cc | cc}
\hline
 &  & \multicolumn{2}{c|}{\textbf{Morph}} & \multicolumn{2}{c}{\textbf{NER}} \\
\textbf{Model} & \textbf{Tok} & \textbf{In} & \textbf{Out} & \textbf{In} & \textbf{Out} \\
\midrule
Baseline             & ULM & 93.04 & 77.73 & 85.53 & 32.17 \\
MorphSeed            & ULM & 92.83 & 78.16 & 84.62 & 37.61 \\
MorphPreTok Actx     & ULM & 93.81 & 81.61 & \textbf{88.69} & \textbf{45.40} \\
MorphPreTok Ctx      & ULM & 93.72 & 82.09 & 88.01 & 42.31 \\
\hline
Baseline             & WP  & 93.09 & 78.94 & 84.22 & 32.94 \\
MorphSeed            & WP  & 93.09 & 78.82 & 85.13 & 35.71 \\
MorphPreTok Actx     & WP  & \textbf{94.04} & 81.68 & 87.22 & 36.82 \\
MorphPreTok Ctx      & WP  & 93.97 & \textbf{82.25} & 88.27 & 44.12 \\
\hline
\end{tabular}
\caption{\textbf{In-domain vs out-domain} whole-string morphological accuracy (Morph) and BI label F1 (NER)}
\label{tab:results_morph_ner_inout}
\end{table} In Table \ref{tab:results_morph_ner_inout}, we observe that increased morphological guidance resulted in the most significant gains for out-of-domain texts. The MorphPreTok variants showed the most improvement, with up to +4.4 morph accuracy and +13.2 BI f1 on out-of domain texts. For NER, in-domain performance was also improved, although there was no difference in in-domain performance for morphological feature classification. 

MorphSeeding did not have a significant effect on morphological feature classification, but for NER had a +5.4 and +2.8 BI F1 gain over baseline for the ULM and WP models, respectively. These results speak to the improved generalization abilities of the morphologically guided tokenizers, especially those with morpheme-based pretokenization (MorphPreTok).

\paragraph{Gains for particular feature values and entities}

\begin{table*}[h]
    \centering
    \scriptsize 
    \begin{tabular}{lc|ccccccccc|cccc}
        \toprule
        & & & \multicolumn{8}{c}{\textbf{Per-Feature Macro F1}} & \multicolumn{3}{c}{\textbf{Per-Entity Micro F1}} \\
        \cmidrule(lr){4-11} \cmidrule(lr){12-14}
        \textbf{Tokenizer Type} & \textbf{Model Class} & \textbf{POS Acc} & \textbf{Case} & \textbf{Degree} & \textbf{Gender} & \textbf{Mood} & \textbf{Number} & \textbf{Person} & \textbf{Tense} & \textbf{Voice} & \textbf{PERS} & \textbf{LOC} & \textbf{GRP} \\
        \midrule
        Baseline & ULM & 94.78 & 72.31 & 90.47 & 92.41 & 79.74 & 95.94 & 97.27 & 90.27 & 95.04 & 65.59 & 60.84 & 67.62 \\
        MorphSeed & ULM & +0.02 & -0.17 & -0.44 & +0.07 & +0.25 & +0.05 & -0.04 & +0.06 & -0.05 & +2.52 & +0.79 & -2.37 \\
        MorphPreTok Actx & ULM & +0.57 & +2.04 & +4.21 & +1.05 & +3.17 & +0.76 & +0.86 & +3.48 & +1.68 & +8.88 & +6.09 & +2.68 \\
        MorphPreTok Ctx & ULM & +0.57 & +6.04 & +4.57 & +1.16 & +3.70 & +0.80 & +0.84 & +3.51 & +1.80 & +7.01 & +4.20 & +1.66 \\
        \midrule
        Baseline & WP & 94.99 & 73.10 & 91.69 & 92.63 & 82.07 & 96.15 & 97.65 & 91.99 & 95.83 & 66.83 & 56.28 & 63.87 \\
        MorphSeed & WP & +0.02 & +4.16 & +0.20 & +0.05 & +0.28 & -0.05 & +0.07 & +0.14 & +0.08 & +0.62 & +4.42 & +4.30 \\
        MorphPreTok Actx & WP & +0.55 & +0.55 & +2.86 & +0.95 & +1.14 & +0.69 & +0.49 & +2.04 & +0.92 & +1.04 & +8.62 & +7.84 \\
        MorphPreTok Ctx & WP & +0.41 & +5.10 & +3.30 & +0.91 & +1.49 & +0.59 & +0.55 & +2.54 & +1.09 & +5.91 & +11.44 & +8.17 \\
        \bottomrule
    \end{tabular}
    \caption{Downstream POS accuracy and per-feature macro F1 scores (POS and Morphological Feature Tagging), and per-entity micro F1 scores (NER). Performance is shown for each baseline and, for morphologically-aware variants, the difference from that class’s baseline.}
    \label{tab:per_feature_macro_f1}
    \vspace{-2mm}
\end{table*}
Table \ref{tab:per_feature_macro_f1} shows the differences in performance compared to baseline tokenizers on particular morphological features and named entities. Again, MorphPreTok is helpful for all features and entities, for both ULM and WP. We see the strongest improvements (+2-6) in per-feature macro-f1 for Case, Degree, Mood, and Tense. Improvements for entities range from +1.66-11.44 in micro f1.

MorphSeed is also helpful for most features and entities, but to a lesser degree than MorphPreTok. Some features see minor regressions (ULM: Case, Degree, Person, Voice; WP: number) and the GRP entity has a more significant -2.37 micro f1 regression for ULM.

\section{Discussion}
The improvements observed in morphological feature tagging and NER are not unexpected. In Latin, morphological features are marked by word endings. Separating these endings from roots allows the model to treat them as informative subunits. Generalization to unseen words is made easier, as their meaning is irrelevant to the task, and their inflectional endings will have been seen during training.

For NER, separating inflectional endings helps the model generalize across different inflected forms of the same entity. Certain entity types, such as locations, also carry morphological markers. For example, the Locative case is used to express “at [place name]”.

Mixed results for sentence-level tasks are also understandable. In word sense disambiguation, the goal is to infer meaning rather than grammatical form, so subwords that encode only inflectional information without independent semantic content may introduce unnecessary noise.

In theory, authorship verification could benefit from morphologically informed tokenization, if for example authors differ in their use of grammatical constructions. While morphologically guided ULM models showed slight gains (62.2 to 67.2 F1) over their baselines, they performed comparably to the baseline WordPiece model (66.8 F1). This suggests that other features, such as lexical choice, may be more informative for this task and could be overshadowed by strict morpheme-based pretokenization.

Although morpheme-based pretokenization improved token-level tasks, we found minimal differences between contextual and acontextual variants. Given the added complexity of contextual pretokenization, its application to other languages may not be justified.

\section{Conclusion}
We demonstrate that morphologically-guided tokenization improves downstream performance in Latin RoBERTa models, particularly for features that are strongly tied to morphological structure. 

More broadly, our results 
highlight the need for continued investment in developing high-quality linguistic resources, particularly for lower-resource and morphologically complex languages, where data availability remains a key bottleneck. 

\section{Limitations} \label{sec:limitations}
Our ULM tokenizers are trained on 5\% of our pretraining corpus, whereas the WordPiece tokenizers are trained on the full dataset. When ULM tokenizers were trained on the full corpus, we observed pathological behavior, including high fertility and segmentations with many single-character subwords and low morphological alignment. Training on smaller datasets, and this 5\% sample, yielded much more regular results, for reasons unclear to us; future work could examine if it is an issue with the HuggingFace implementation. We decided to train ULM tokenizers on a subset of the corpus, in order to have higher-quality tokenization and a fairer comparison to the WordPiece tokenizers. All RoBERTa models were pretrained on the full corpus.

We only experiment with encoder-based models. The performance gains we observed may not scale to larger models or other architecture types.

We only pretrain and finetune a single model per tokenizer, in order to reduce computational time and cost. 


\section{Acknowledgments}
We would like to thank EleutherAI for providing computational resources used for this paper. Additionally, we thank the UMass NLP group and Catherine Arnett for their support and feedback.
This material is based in part upon work
supported by National Science Foundation award
1845576 (CAREER). Any opinions, findings and conclusions
or recommendations expressed in this material are
those of the authors and do not necessarily reflect
the views of the National Science Foundation.

\bibliography{anthology,custom}

\appendix

\section{Appendix}\label{sec:appendix}

\begin{table}[h]
    \centering
    \setlength{\tabcolsep}{2pt}
    \scriptsize 
    \begin{tabular}{@{}ll|ccccc@{}}
        \toprule
        & & \multicolumn{5}{c}{\textbf{Gold = Actx. Sigmorphon Segmentations}} \\
        \textbf{Tokenizer Type} & \textbf{Alg} & \textbf{EM} & \textbf{Recall} & \textbf{Precision} & \textbf{F1} & \textbf{Fertility} \\
        \midrule
        Baseline & ULM & 7.19 & 20.30 & 16.33 & 18.10 & 3.0964 \\
        MorphSeed & ULM & 7.03 & 20.72 & 16.69 & 18.49 & 3.0907 \\
        MorphPreTok Actx & ULM & 9.89 & 26.23 & 20.50 & 23.01 & 3.1904 \\
        MorphPreTok Ctx & ULM & 9.34 & 25.63 & 19.99 & 22.46 & 3.2001 \\ \hline
        Baseline & WP  & 3.60 & 12.29 & 10.77 & 11.48 & 2.8434 \\
        MorphSeed & WP  & 3.46 & 11.56 & 10.43 & 10.97 & 2.7615 \\
        MorphPreTok Actx & WP  & 8.45 & 19.17 & 17.93 & 18.53 & 2.6651 \\
        MorphPreTok Ctx & WP  & 8.28 & 18.94 & 17.83 & 18.37 & 2.6482 \\
        \midrule
        & & \multicolumn{5}{c}{\textbf{Gold = Ctx. Lemlat Segmentations}} \\
        \textbf{Tokenizer Type} & \textbf{Alg} & \textbf{EM} & \textbf{Recall} & \textbf{Precision} & \textbf{F1} & \textbf{Fertility} \\
        \midrule
        Baseline & ULM & 10.76 & 11.68 & 12.08 & 11.88 & 1.8714 \\
        MorphSeed & ULM & 12.12 & 13.72 & 14.04 & 13.88 & 1.8926 \\
        MorphPreTok Actx & ULM & 65.05 & 77.62 & 63.74 & 70.00 & 2.3585 \\
        MorphPreTok Ctx & ULM & 71.88 & 84.98 & 68.43 & 75.81 & 2.4052 \\ \hline
        Baseline & WP  & 20.12 & 21.37 & 23.41 & 22.34 & 1.7688 \\
        MorphSeed & WP  & 20.18 & 21.49 & 23.53 & 22.46 & 1.7688 \\
        MorphPreTok Actx& WP  & 75.50 & 82.94 & 75.30 & 78.93 & 2.1335 \\
        MorphPreTok Ctx & WP  & 84.32 & 91.90 & 81.82 & 86.57 & 2.1755 \\
        \bottomrule
    \end{tabular}
    \caption{Full tokenizer evaluation metrics across acontextual (Sigmorphon, top) and contextual (Lemlat, bottom) gold segmentations. Recall, precision, and f1 are in terms of subword overlap between the predicted and gold segmentations.}
    \label{tab:tokenizer_eval}
\end{table}

\subsection{Disambiguating Segmentations with POS Tags}
\begin{table}[h]
    \centering
    \begin{tabular}{ll}
        \toprule
        \textbf{UD POS Tag} & \textbf{Lemlat POS Tags} \\
        \midrule
        NOUN & Noun, Adjective \\
        PROPN & Noun, Adjective \\
        VERB & Verb \\
        ADJ & Adjective, Noun \\
        PRON & Pronoun, Noun, Invariable \\
        ADV & Invariable \\
        ADP & Preposition, Invariable \\
        CCONJ & Conjunction, Invariable \\
        SCONJ & Conjunction, Invariable \\
        PART & Interjection, Invariable \\
        INTJ & Interjection, Invariable \\
        DET & Pronoun, Adjective \\
        X & Invariable, Other \\
        AUX & Verb \\
        PUNCT & Invariable \\
        NUM & Noun, Adjective, Invariable \\
        \bottomrule
    \end{tabular}
    \caption{Mapping from Universal Dependencies (UD) POS Tags to Lemlat POS Tags}
    \label{tab:ud_to_lemlat}
\end{table}
When attempting to match a UD POS tag to a Lemlat POS tag, the Lemlat tags are checked in the order they appear in Table \ref{tab:ud_to_lemlat}. 

\subsection{Tokenizer Implementation Details}\label{sec:appendix_tok_details}
\paragraph{Training Hyperparameters} 
For all tokenizers, we fix the vocabulary size at 30k. For ULM, we set the shrinking factor to the default HuggingFace value, 0.75.

\paragraph{MorphPreTokenization}
For ULM, we use a sequence of the default \verb|Metaspace()| pretokenizer, followed by a \verb|CharDelimiterSplit(delimiter="@")|. The \verb|Metaspace| pretokenizer replaces whitespace with a special underscore-like symbol, 
then splits on this character and prepends it to the next word. Functionally, this means that the first subword of a word is differentiated from continuing subwords with this symbol. Then, the \verb|CharDelimiterSplit| pretokenizer will split on our morpheme symbol, allowing morphological suffixes to be treated as continuing subwords. 

WordPiece requires more modification than ULM, since the continuing subword symbol "\#\#" is added at train time rather during pretokenization. First, we use a sequence of pretokenizers: \verb|WhitespaceSplit()| followed by \texttt{Split(pattern = "@", behavior = "merged\_with\_next")}. Then, we add a \verb|morph_delimiter="@"| argument to the \verb|WordPieceTrainer|. During training, if a subword is encountered that starts with the \verb|morph_delimiter|, the delimiter is replaced with the continuing subword symbol "\#\#" and the new subword is added to the vocabulary.

\subsection{Pretraining Details} \label{sec:appendix_pretrain_details}

\begin{table}[h!]
    \centering
    \small
    \setlength{\tabcolsep}{6pt}
        \begin{tabular}{l l}
        \toprule
        \textbf{Hyperparameter} & \textbf{Value} \\
        \midrule
        Epochs & 10 \\
        Effective batch size & 256 \\
        Learning rate & $2 \times 10^{-5}$ \\
        Weight decay & 0.01 \\
        Warmup steps & 1000 \\
        Max position embeddings & 513 \\
        Number of attention heads & 12 \\
        Number of hidden layers & 12 \\
        Hidden size & 768 \\
        MLM probability & 0.15 \\
        Architecture & \texttt{RobertaPreLayerNorm} \\
        \bottomrule
        \end{tabular}
    \caption{Pretraining hyperparameters for all 8 models. Training follows a HuggingFace \texttt{RobertaPreLayerNorm} architecture with masked language modeling (MLM) objective.}
    \label{tab:pretraining_hyperparams}
\end{table}

For all 8 models, pretraining took around 4460 GPU hours on A40s and H200s.

\subsection{Finetuning Details} \label{sec:appendix_finetune_details}

\subsubsection{Hyperparameters}
\begin{table}[h!]
    \centering
    \small
    \setlength{\tabcolsep}{4pt}
        \begin{tabular}{l | l l l}
        \toprule
        \textbf{Hyperparameter} & \textbf{POS/Morph} & \textbf{NER} & \textbf{WSD} \\
        \midrule
        Batch size & 8 & 16 & 8 \\
        Epochs & 15 & 10 & 20 \\
        Dropout & 0.1 & -- & 0.25 \\
        Weight decay & 0.01 & 0.01 & -- \\
        Learning rate & $5 \times 10^{-5}$ & $2 \times 10^{-5}$ & $5 \times 10^{-5}$ \\
        \bottomrule
        \end{tabular}
    \caption{Fine-tuning hyperparameters for downstream tasks.}
    \label{tab:finetune_hyperparams}
\end{table}
Hyperparameters are taken from prior work in Latin POS/Morphological feature tagging \cite{hudspeth-etal-2024-latin}, NER \cite{beersmans-etal-2023-training}, WSD \cite{ghinassi-etal-2024-language}.

For WSD, we specicially train on \citet{ghinassi-etal-2024-language}'s SemEval + Pers$_{inter}$ train set.

To determine which checkpoint is best-performing, we use the validation whole-string morphological accuracy for POS/Morphological feature tagging, and the validation F1 for NER. WSD always saves the model from the last epoch.

The AV task does not require finetuning. We re-implemented the baseline method described by \citet{gorovaia-etal-2024-sui}.

\subsubsection{Subword Aggregation Strategies} 
\begin{table}[h!]
\centering
\small
\setlength{\tabcolsep}{4pt}
    \begin{tabular}{l l | c c c}
    \toprule
     &  & \multicolumn{3}{c}{\textbf{Morph Acc}} \\
    \textbf{Model} & \textbf{Tok} & \textbf{First} & \textbf{Last} & \textbf{Mean} \\
    \midrule
    Baseline & ULM & 76.88 & 77.73 & 75.18 \\
    MorphSeed & ULM & 77.44 & 78.16 & 75.64 \\
    MorphPreTok Actx & ULM & 81.76 & 81.61 & 78.58 \\
    MorphPreTok Ctx & ULM & 81.70 & 82.09 & 78.51 \\
    \hline
    Baseline & WP & 77.19 & 78.94 & 77.36 \\
    MorphSeed & WP & 77.40 & 78.82 & 77.33 \\
    MorphPreTok Actx & WP & 81.11 & 81.68 & 78.96 \\
    MorphPreTok Ctx & WP & 81.41 & \textbf{82.25} & 79.54 \\
    \midrule
    \textit{Per-col AVG} &  & 79.36 & \textbf{80.16} & 77.64 \\
    \midrule
     &  & \multicolumn{3}{c}{\textbf{NER BI F1}} \\
     \textbf{Model} & \textbf{Tok} & \textbf{First} & \textbf{Last} & \textbf{Mean} \\
    \midrule
    Baseline & ULM & 28.91 & 32.17 & 29.56 \\
    MorphSeed & ULM & 30.15 & 37.61 & 31.04 \\
    MorphPreTok Actx & ULM & 44.47 & 45.40 & \textbf{46.40} \\
    MorphPreTok Ctx & ULM & 42.70 & 42.31 & 43.99 \\
    \hline
    Baseline & WP & 30.22 & 32.94 & 30.96 \\
    MorphSeed & WP & 34.76 & 35.71 & 33.25 \\
    MorphPreTok Actx & WP & 36.81 & 36.82 & 41.11 \\
    MorphPreTok Ctx & WP & 36.01 & 44.12 & 39.33 \\
    \midrule
    \textit{Per-col AVG} &  & 35.50 & \textbf{38.39} & 36.96 \\
    \bottomrule
    \end{tabular}
\caption{Out-domain performance across \textbf{subword aggregation strategies} (first subword, last subword, mean pooling) for Morph and NER tasks.}
\label{tab:results_pool_strategy_comparison}
\end{table}

As reported in Table \ref{tab:results_pool_strategy_comparison}, we generally saw the best performance on the token level tasks when predicting from the last subword. For Latin, we did not find any prior literature using the last subword, but we did find prior work using the first subword embedding \cite{riemenschneider-frank-2023-exploring} and the mean-pooled word embedding \cite{bamman2020latinbertcontextuallanguage}. 

\subsection{Downstream Task Descriptions}
\label{sec:app_task_desc}
\begin{table}[h!]
    \centering
    \small
    \begin{tabular}{l | r r l}
        \hline
        \textbf{Task} & \textbf{In-domain} & \textbf{Out-domain} & \textbf{Unit} \\
        \midrule
        Morph & 68{,}103 & 24{,}414 & words \\
        NER & 1{,}048 & 570 & BI labels \\
        \hline
    \end{tabular}
    \caption{Sizes of in and out domain test sets for POS/Morphological Feature tagging and NER.}
    \label{tab:testset_sizes_inout}
\end{table}

\subsubsection{POS and Morphological Feature Tagging}
\paragraph{Data} We use the official train/test splits of five Latin Universal Dependencies (UD) Treebanks \cite{de-marneffe-etal-2021-universal},\footnote{Perseus \cite{perseus}, PROIEL \cite{proiel}, LLCT \cite{llct}, ITTB \cite{ittb}, and UDante \cite{udante}: \url{https://universaldependencies.org/la/}}
harmonized to have more consistent morphological features \cite{gamba-zeman-2023-latin} and standardized to use Latin-specific morphological features \cite{hudspeth-etal-2024-latin}.

When comparing in versus out domain performance, we consider the Perseus and UDante test sets as the out-domain. They comprise the smallest portion of the finetuning data, and they are stylistically distinct from the other treebanks: Perseus is primarily Classical era histories, poems, epics, satires; and UDante Medieval treatises, letters, poems \cite{hudspeth-etal-2024-latin}.

\paragraph{Modeling}
We use a separate classification head for each morphological feature, the same architecture as \citet{riemenschneider-frank-2023-exploring}'s finetuned Greek model. See \S\ref{sec:appendix_finetune_details} for details on hyperparameters. Additionally, we experiment with three subword aggregation strategies, as we noticed different methods being used in prior work: predicting from the first subword (as in \citet{riemenschneider-frank-2023-exploring}), predicting from the last subword, and predicting from the averaged word embedding (as in \citet{bamman2020latinbertcontextuallanguage}'s POS tagging task for LatinBERT).

\paragraph{Metrics}
We report whole-string morphological accuracy, following the convention of \citet{gamba-zeman-2023-latin} and \citet{sprugnoli-etal-2022-overview}. This metric considers the model’s prediction correct when every morphological feature is correctly predicted, indicating whether the model understands how all the morphological features fit together. 

Additionally, we report per-feature macro-F1 scores in order to emphasize the performance on rare feature values.

These metrics are aggregated across the test sets, weighted by the number of labels (words) in each test treebank.

\subsubsection{Named Entity Recognition (NER)} 
\paragraph{Data} We finetune our models on the Herodotos Project's\footnote{\url{https://u.osu.edu/herodotos/}} manually annotated NER dataset \cite{erdmann-etal-2016-challenges,erdmann-etal-2019-practical}, using \citet{beersmans-etal-2023-training}'s in-domain/out-domain splits. The in-domain consists of prose texts (Caesar’s \textit{Bellum Gallicum} and \textit{Bellum Civile}, Pliny the Younger’s \textit{Epistulae}, and Pliny
the Elder’s \textit{Naturalis Historia}), and the out-domain a single poetry text, Ovid’s \textit{Ars
Amatoria}. 


\paragraph{Modeling} Following \citet{beersmans-etal-2023-training}'s finetuning of LatinBERT, we treat this as a single-head token classification task. 
Similarly to our morphological feature classification task, we test three subword aggregation strategies: for each word, predicting on the first subword, the last subword, or an average embedding of that word's subwords. 

\paragraph{Metrics} We primarily report the overall micro f1 (accuracy) for BI labels, excluding the O label. 

\subsubsection{Word Sense Disambiguation (WSD)}

\paragraph{Data} 
We employ \citet{ghinassi-etal-2024-language}'s train/test split of SemEval's 2020 shared task on
Unsupervised Lexical Semantic Change Detection \cite{schlechtweg-etal-2020-semeval}. In addition to the SemEval training data, we also train on \citet{ghinassi-etal-2024-language}'s Pers$_{inter}$ silver data, created by propagating word senses from English. Unlike previous Latin WSD experiments \cite{bamman2020latinbertcontextuallanguage,lendvai-wick-2022-finetuning}, this dataset includes more than two senses per lemma, derives senses from multiple dictionaries, and includes texts from a wider time range.

\paragraph{Modeling} We replicate \citet{ghinassi-etal-2024-language}, who finetuned LatinBERT using a single classification head on top of a mean-pooled representation of the target word's subword embeddings. 

\paragraph{Metric} We report the average F1 across the 40 lemmas in the test set.

\subsubsection{Authorship Verification}

\paragraph{Data} The dataset is a subset of the Patristic
Sermon Textual Archive (PaSTA), curated by \citet{gorovaia-etal-2024-sui}. They selected 22 authors who preached during the 3rd to 7th centuries, sampling 15 positive and 15 negative text pairs per author. In total, there are 660 text pairs.

\paragraph{Method} We replicate \citet{gorovaia-etal-2024-sui}'s baseline method, which generates sentence embeddings from the base model using mean-pooling, tunes a cosine similarity threshold on two-thirds of the data, and tests the threshold on the remaining one-third of the data. Unlike the other downstream tasks, this method does not require finetuning.

\paragraph{Metrics} We divide the data into three disjoint portions and conduct three trials, each using a different portion as the test set, and report the average F1 score across trials. Thus, the size of the test set will always be 220 text pairs.

\subsection{Effect of Word Frequency on Morphological Classification Accuracy}
\begin{figure}[h]
    \vspace{-2mm}
    \centering
    \begin{minipage}{\linewidth}
        \centering
        \includegraphics[width=\linewidth]{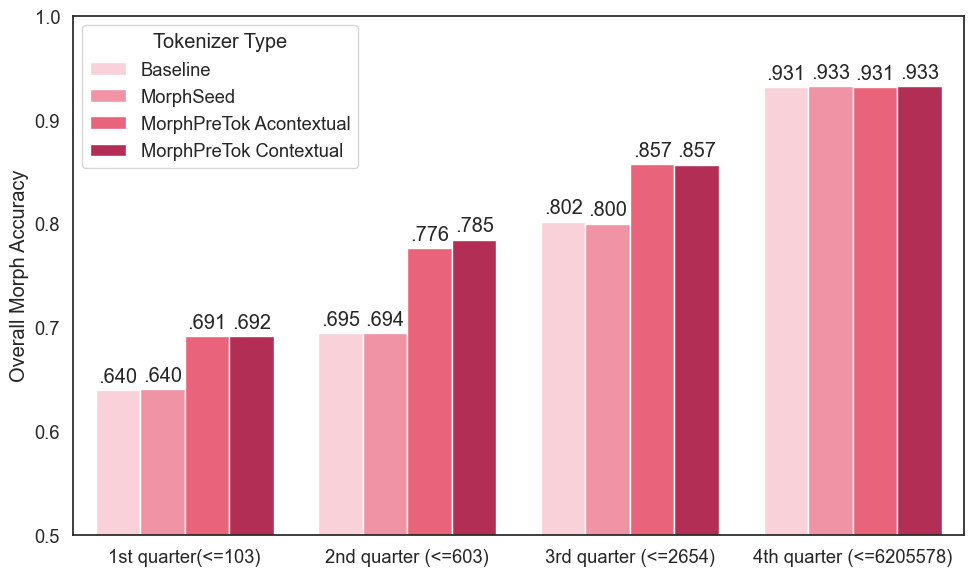}
        \label{fig:freq_ulm}
    \end{minipage}
    
    \vspace{-6mm} 

    \begin{minipage}{\linewidth}
        \centering
        \includegraphics[width=\linewidth]{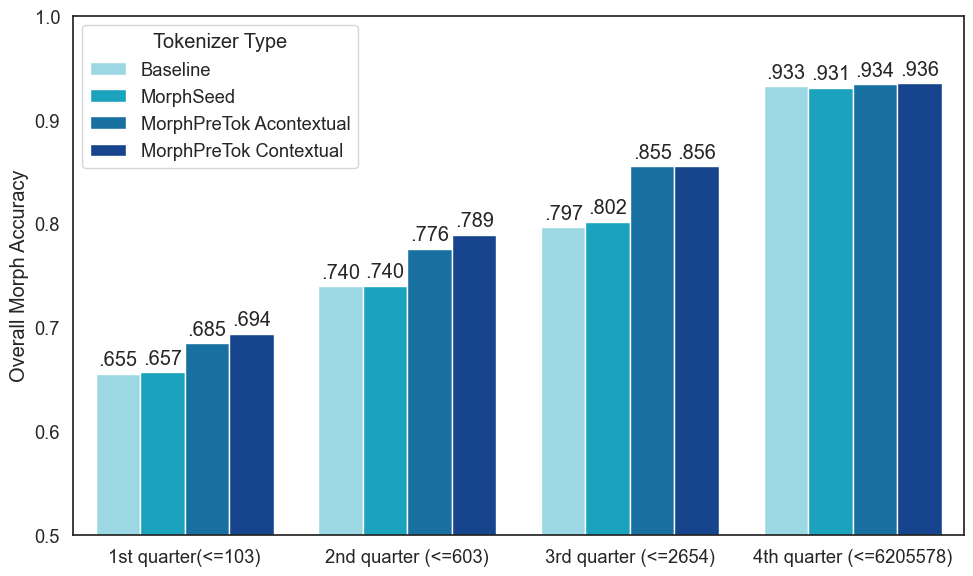}
        \label{fig:freq_wp}
    \end{minipage}
    \vspace{-6mm}
    \caption{Word frequency in the pretraining corpus versus whole-string morphological accuracy, for ULM (top) and WordPiece (bottom).}
    \label{fig:freq_comparison}
    \vspace{-2mm}
\end{figure}

Performance on morphological feature classification is unchanged for the words seen most frequently in the pretraining corpus, but rarer words see a boost from the MorphPreTok tokenizers, indicating better generalization.

\subsection{Performance of Existing Models on Downstream Tasks}
\label{sec:app_existing_models}

\begin{table}[h!]
    \centering
    \small
    
    \begin{tabular}{l l | c c c c}
        \hline
        \textbf{Model} & \textbf{Tok} & \textbf{Morph} & \textbf{NER} & \textbf{WSD} & \textbf{AV} \\
        \midrule
        LatinBERT & WP & \textbf{96.07} & 81.19* & \textbf{61.45*} & \textbf{72.60} \\
        LaBERTa & BPE & 88.86 & 77.60/\textbf{85.21} & 52.20 & 70.70 \\
        \hline
    \end{tabular}
    \caption{Summary of results on downstream tasks for existing Latin encoder models. For Morph and NER, reported results use mean pooled word representations for prediction. For LaBERTa NER, we show results on lowercased/cased data, since LaBERTa's tokenizer is cased but LatinBERT's is not. LatinBERT's WSD* f1 taken from \citet{ghinassi-etal-2024-language}, and NER* BI f1 computed based on reported in and out domain BI f1 in \citet{beersmans-etal-2023-training}.}
    \label{tab:existing_models}
\end{table}

LatinBERT dominates on most downstream tasks, likely because its pretraining corpus (643M words) is nearly 4x larger than LaBERTa's (165M) or the corpus used 
in this work (\S\ref{sec:datamodels}).

\end{document}